\documentclass{amos}

\title{\TitleFont Closely-Spaced Object Classification Using MuyGPyS}

\author{Kerianne Pruett \\ Lawrence Livermore National Laboratory \and \vspace{.9em} Nathan McNaughton \\ University of California, Berekeley \and Michael Schneider \\ Lawrence Livermore National Laboratory}

\date{}

\begin{document} 

\maketitle

\begin{abstract}
	\normalsize
	Accurately detecting rendezvous and proximity operations (RPO) is crucial for understanding how objects are behaving in the space domain. However, detecting closely-spaced objects (CSO) is challenging for ground-based optical space domain awareness (SDA) algorithms as two objects close together along the line-of-sight can appear blended as a single object within the point-spread function (PSF) of the optical system. Traditional machine learning methods can be useful for differentiating between singular objects and closely-spaced objects, but many methods require large training sample sizes or high signal-to-noise conditions. The quality and quantity of realistic data make probabilistic classification methods a superior approach, as they are better suited to handle these data inadequacies. We present CSO classification results using the Gaussian process python package, \texttt{MuyGPyS}, and examine classification accuracy as a function of angular separation and magnitude difference between the simulated satellites. This orbit-independent analysis is done on highly accurate simulated SDA images that emulate realistic ground-based commercial-of-the-shelf (COTS) optical sensor observations of CSOs. We find that \texttt{MuyGPyS} outperforms traditional machine learning methods, especially under more challenging circumstances. 
	
\end{abstract}

\section{Introduction}
	\label{sec:intro}

	The safety and security of space missions requires a firm understanding of how objects are behaving and operating within the space domain, making accurate detection methods a crucial capability. Detecting rendezvous and proximity operation (RPO) maneuvers challenges traditional ground-based optical space domain awareness (SDA) methods, as two objects close together can fall within the point-spread function (PSF) of the optical system, thus appearing as a single object. 
	
	Machine learning applications can be utilized to disambiguate closely-spaced objects (CSO) in images, however, most traditional methods rely on a large number of training samples, or classification accuracy begins to break down rapidly in low signal-to-noise regimes. In contrast, Gaussian processes enable one to take a probabilistic approach to the classification problem, requiring less training samples and performing better in the presence of noisy data. Gaussian processes additionally produce probabilistic outputs which enable reliable flagging of ambiguous images for follow-up with human inspection. RPO data exists in limited quantities, and optical images are often of limited quality due to seeing, weather, or sensor parameters, making Gaussian process approaches better suited for this classification task than in comparison to traditional machine learning methods. 
	
	We present our CSO classification results using \texttt{MuyGPyS}\footnote{https://github.com/LLNL/MuyGPyS}, a fast and flexible Gaussian Process (GP) python package developed by Lawrence Livermore National Laboratory (LLNL) \cite{Muyskens:2021}. We compare the MuyGPS methods' ability to detect CSOs as a function of angular distance, magnitude difference between objects, and across different signal-to-noise regimes. Conducting the analysis in this way allows for results that are independent of orbital regime or optical sensor specifications, as the optical effects of sensor performance and satellite altitude fall within the sampled parameter spaces. 
	
	Images used in this analysis are simulated using LLNL's image simulation suite, which utilizes \texttt{GalSim}\footnote{https://github.com/GalSim-developers/GalSim} \cite{Rowe:2015}, an open-source python package for simulating highly accurate ground-based optical images. Satellites are then injected into \texttt{GalSim} images using the LLNL-developed orbital dynamics simulation package for space situational awareness, \texttt{SSAPy}. We assume Lambertian sphere emission models and commercial-off-the-shelf (COTS) ground-based optical sensors, and our image simulations include accurate atmosphere and noise models given various observing conditions. We find that our Gaussian process approach outperforms conventional machine learning methods, especially in the more challenging regions of simulation parameter space.

\section{Modeling Tools}

LLNL has developed a suite of SDA modeling tools in-house, include orbital dynamics packages, SDA image simulation suites, and Gaussian Process machine learning methods that can be used and applied to challenging SDA image classification tasks. Below is a brief overview of each package used in this work. All of our packages have been adapted for high-performance computing use, and have been used to carry out extensive, large-scale simulations.

\subsection{SSAPy}

\texttt{SSAPy} is LLNL's python package for space situational awareness (SSA). \texttt{SSAPy} is a fast, flexible, and numerically stable orbital dynamics package, that allows the users to define and propagate orbits given a multitude of user-specified parameters.

Orbits can be defined in various ways, such as reading in data from a TLE file, specifying orbital elements (Keplerian, Equinoctial, Kozai Mean, etc.). Users can customize a force propagation model by including accelerations parameters such as the accelerations from the Sun, Moon, all of the planets, harmonic accerlarions (from the Earth and Lunar surfaces, with many gravity model options to choose from), Earth radiation pressure, solar radiation pressure, atmospheric drag, Keplerian accelerations, etc. \texttt{SSAPy} supports multiple integrators such as Runge-Kutta (4/5 and 7/8), \texttt{Scipy} integrators, SGP4, Keplerian, and Taylor Series. Users may define maneuvers by numerically integrating over arbitrary thrust profiles, assuming constant acceleration between each user defined time step. \texttt{SSAPy} also includes many built-in functions for coordinate frame conversions, data parsing, and plotting. The user can return information for a constructed orbit at any timestep in that orbit, such as the state vector, specific angular momentum, specific orbital energy, any orbital element, the period, periapsis, and apoapsis of the orbit, and can even generate a TLE at any given time.

\texttt{SSAPy} supports Monte Carlo appplications for probabalistic modeling, and has been used for applications such as target tracking data fusion with distributed sensors \cite{Miller:2020}, short-arc probabalistic orbit determination, estimating rare conjunction probabilities \cite{Miller:2022}, and for large-scale cislunar orbital stability simulations through the generation of millions of cislunar orbits over long time scales (Yeager, T., Pruett, K., and Schneider, M. in prep). 

\subsection{Image Simulation Suite}

LLNL's SDA image simulation suite includes packages for simulating realistic ground-based electro-optical (EO) images. The simulation suite utilizes \texttt{SSAPy} and the open-source EO image simulation package \texttt{GalSim}. \texttt{GalSim} gives the user control over various parameters, such as detector parameters, optical system parameters, and realistic atmospheric and photometric effects. This wrapper for \texttt{GalSim} enables us to simulate images for any ground-based sensor and satellite pair. Our package includes support for modifying the vignetting, optical distortion, and observing conditions, and also include real star positions and fluxes utilizing stars and data from the Gaia\footnote{https://www.esa.int/Science\_Exploration/Space\_Science/Gaia\_overview} catalog. 

\begin{figure}[h!]
	\centering
	\includegraphics[width=0.95\textwidth]{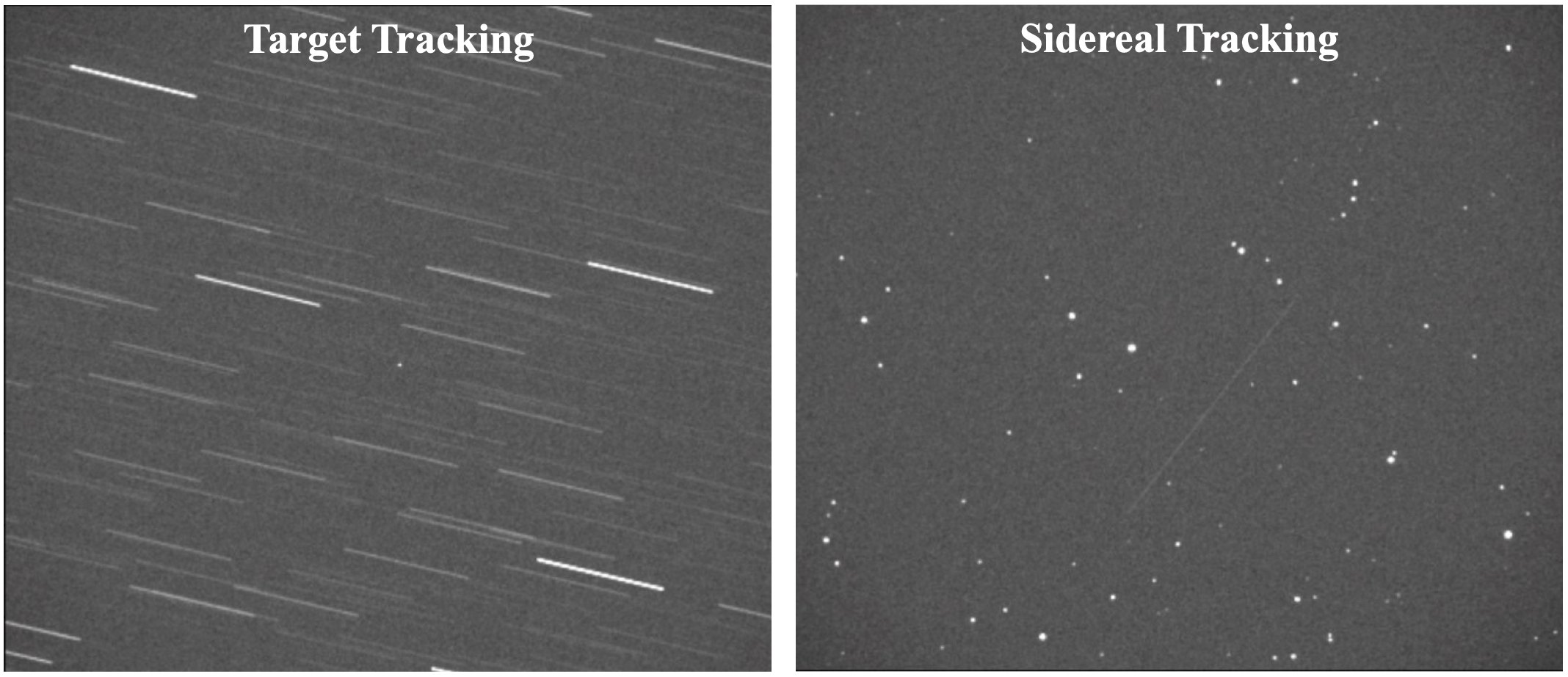}
	\caption{Simulated SDA images in different tracking modes.}
	\label{fig:xfiles}
\end{figure}

Parameters can be set manually given real desired sensor parameters, determined using real data, or can be set through ray tracing over the optical system (i.e. using open-source packages such as Batoid\footnote{https://github.com/jmeyers314/batoid}). For example, if the user knows the sensor and satellite (size, orbit, etc.) they want to simulate, they can put in those parameters to generate a realistic image, or, if the user wants to quickly simulate an image given an SNR, sky background, and satellite magnitude, they can skip to inputting those values instead of setting up the realistic parameters. The addition of the ray tracing option additionally allows a user to simulate images for a system they don't have yet, or want to understand the capabilities of. Our models currently assume a Lambertian Sphere model with spherically-symmetric emissivity. Our simulation packages support target tracking mode (sensor tracking the satellite during the exposure) and sidereal tracking mode (tracking the fixed stars, and not moving over the exposure), and a user can specify to request multiple exposures of a satellite throughout it's track, or plug-in to \texttt{SSAPy} for simulating maneuvers between exposures, etc. A full-sized example of a target tracking and sidereal tracking image with our software suite can be found in Fig. \ref{fig:xfiles} (in the left panel the satellite is a point near the center of the image, and in the left panel the satellite is a streak, just southeast of the center).

\subsection{MuyGPyS}

\texttt{MuyGPyS} is the open-source python implementation of the LLNL-developed MuyGPs method \cite{Muyskens:2021}. \texttt{MuyGPyS} is a fast and flexible Gaussian Process hyperparamter estimation method that has been demonstrated to be faster and more accurate in comparison to other state-of-the-art GP methods \cite{Muyskens:2021, Muyskens:2022}. \texttt{MuyGPyS} has been applied to SDA-related classification purposes such as COTS light curve completion \cite{Goumiri:2022}, star vs. galaxy classification \cite{Muyskens:2022, Goumiri:2020}, and galaxy blend classification prediction for the Rubin Observatory's Legacy Survey of Space and Time (LSST)\footnote{https://www.lsst.org/about} \cite{Buchanan:2022}.

\section{Data Simulation}

\subsection{Data Assumptions}

In this work, we use our image simulation suite to simulate ground-based EO images of low-Earth orbit (LEO) satellites in the i-band, in which some images have a single satellite, and others have two satellites (i.e. ``CSOs"). We choose an i-band filter for the simulated images, as this band produced the highest classification accuracy when compared to the u-, g-, r-, and z-bands \cite{Muyskens:2022}. Other relevant parameters that we fix or set for our simulations are listed in Table \ref{tab:simparams}. If there are two values in brackets, this means we sample a float value between the two bounds. If there are more than two values in brackets, this means that we sample one of the discrete values listed there. 

Our simulations rely on ``nominal" COTS optical SDA parameters discussed throughout the literature or supplied to us through members of the SDA community. Conducting our analysis in this way (by selecting values from ranges instead of simulating a particular sensor and satellite pair) enables us to perform an analysis that is independent of site, sensor, and orbital regime independent, as the observing conditions, satellite size, magnitude, and albedo, and other optical sensor effects should all fall within our sampled parameter spaces.

\begin{table}[h!]
	\begin{center}
		\caption{Image Simulation Parameters}
		\begin{tabular}{|c|c|}
			\hline
			\textbf{Parameter} & \textbf{Value} \\ 
			\hline
			Filter & i-band  \\ 
			Primary Satellite Magnitude & [12, 15] \\ 
			Secondary Satellite Magnitude & [-1.5, 1.5] $\times$ Primary Satellite Mag \\
			PSF FWHM & [2, 6] $arcseconds$ \\
			CSO Distance Offset - RA & Primary Satellite RA + ($\pm1 \times$ [0.2, 1.5] $\times$ PSF FWHM) \\
			CSO Distance Offset - DEC & Primary Satellite DEC + ($\pm1$ $\times$ [0.2, 1.5] $\times$ PSF FWHM) \\
			Exposure Time & [0.1, 0.2, 0.5] $seconds$ \\
			Full Image Size & 1800$\times$1536 $pixels$ \\
			Cutout Size & 24$\times$24 $pixels$ \\
			Sky Background Range & [18.3, 20.3] $mag$ $arsec^{-2}$ \\
			Zero-Point Range & [0, .3] $\times$ $log_{10}$(scaled zero point)$/0.4 + 24$\\
			Size Distribution & [.1, 10] $meters$ \\
			Albedo Distribution & [0.05, 0.2] \\
			Pixel Scale & 25.0 $microns$ \\
    		Gain & 1.0 \\
    		Read Noise & 5.0 \\
    		Aperture & 1.0 $meter$ \\
    		Obscuration & 0.1 (fractional linear)\\
			Jitter & 0.00139 $degrees$\\
			\hline
		\end{tabular}
		\label{tab:simparams}
	\end{center}
\end{table}

For the scope of this work, we assume that the ground-based sensor taking the image knows that there is at least one satellite present, and it is tracking that satellite during the sampled exposure time. This satellite we refer to as the ``primary satellite''. Repeating the analysis for sidereal tracking sensor modes would add a detection component to our analysis, but for the scope of this work are concerned with disambiguating between one or two satellites, assuming one satellite is already known to be present. 

In addition to the parameters in Table \ref{tab:simparams}, we simulate realistic vignetting, optical distortion, and sensor tracking erorrs for a ``"nominal SDA COTS optical system. To create the cuouts, a full-sized image is first simulated (such as the left panel of Fig. \ref{fig:xfiles}). Next, the cutout is made by determining where the primary satellite is located within the image (accounting for realistic error in detecting and calibrating the location of the known primary satellite), and removing all the pixels that lie beyond a 24x24 bounding box around the primary satellites calculated position.

\subsection{Data Pre-Processing}

\begin{figure}[p]
	\centering
	\includegraphics[width=0.95\textwidth]{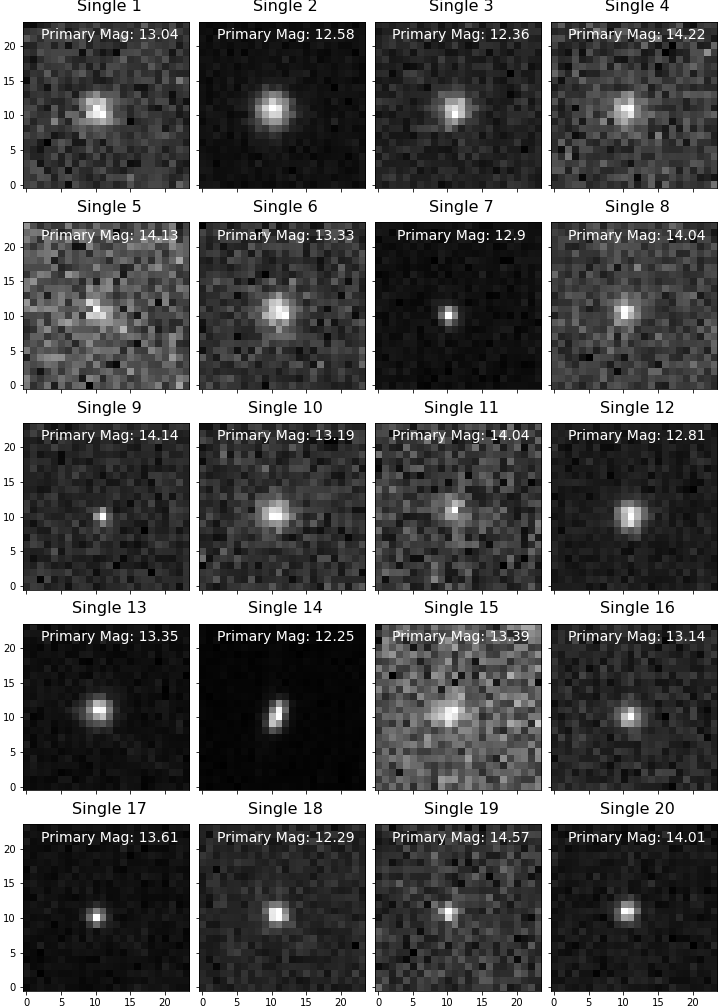}
	\caption{Simulated cutouts of singular satellites, simulated in the i-band.}
	\label{fig:singlesats}
\end{figure}

\begin{figure}[p]
	\centering
	\includegraphics[width=0.95\textwidth]{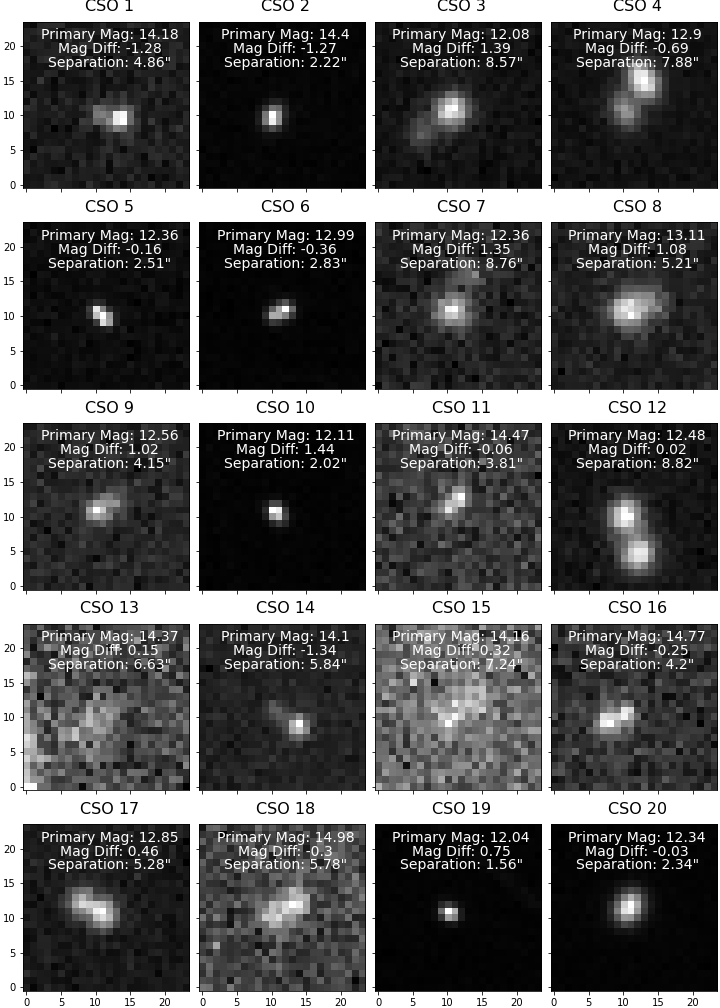}
	\caption{Simulated cutouts with two satellites in close proximity, simulated in the i-band.}
	\label{fig:csosats}
\end{figure}

We tried multiple normalization techniques, including local minmax (i.e. over each image), global minmax (i.e. over all images), log normalization, and the normalization technique described in \cite{Muyskens:2022} (which involves subtracting the minimum pixel value off of each pixel in an image, and dividing by the maximum pixel value accross all images). Of all methods tried, we found the best performance using the simple local minmax scaling method:

\begin{equation}
	{x_{I}}^{'}=\frac{x_{i}-min(x)}{max(x)-min(x)},
\end{equation}

where $x_{i}$ is the current pixel, $x$ is the full array of pixels for a given image, and ${x_{I}}^{'}$ is the new updated normalized pixel at index $i$. Our dataset consists of 6,977 single satellite cutouts, and 4,977 double satellite cutouts (a.k.a. ``CSOs''). Sample cutouts for each of our classes, with normalization techniques applied, are presented in Fig. \ref{fig:singlesats} and Fig. \ref{fig:csosats}.

For CSO cutouts, we calculate the angular distance and magnitude difference between the primary and secondary satellites. To determine the angular distance between a pair, we take the truth information from our simulation and calculate the angular distance between the center of the primary satellite ($RA_p$) and the center of the secondary satellite ($RA_s$). Because the exposures are $>0$ seconds, there are technically endpoints associated with each simulated satellite, but because the exposure times are small enough (i.e. the satellite appears as a circular blob rather than a streak in the image), we simply assume the the satellite is located at the midpoint between the initial location ($RA_i$, $DEC_i$) and final location ($RA_f$, $DEC_f$), where $RA$ and $DEC$ are the right ascension and declination, respectively. To calculate the $RA$ and $DEC$ for each satellite pair we use the following equation: 

\begin{equation}
	(\text{RA}_x, \text{DEC}_x) = \Bigl(\frac{\text{RA}_x^i + \text{RA}_x^f}{2} , \frac{\text{DEC}_x^i + \text{DEC}_x^f}{2}\Bigr)
\end{equation}

where $_x$ can be replaced with $_p$ for the primary satellite and $_s$ for the secondary satellite. Once the two sets of $RA$ and $DEC$ are calculated, we determine the angular separation between the two satellites using the great circle distance between two points on a sphere. 

\begin{figure}[h!]
	\centering
	\includegraphics[width=0.99\textwidth]{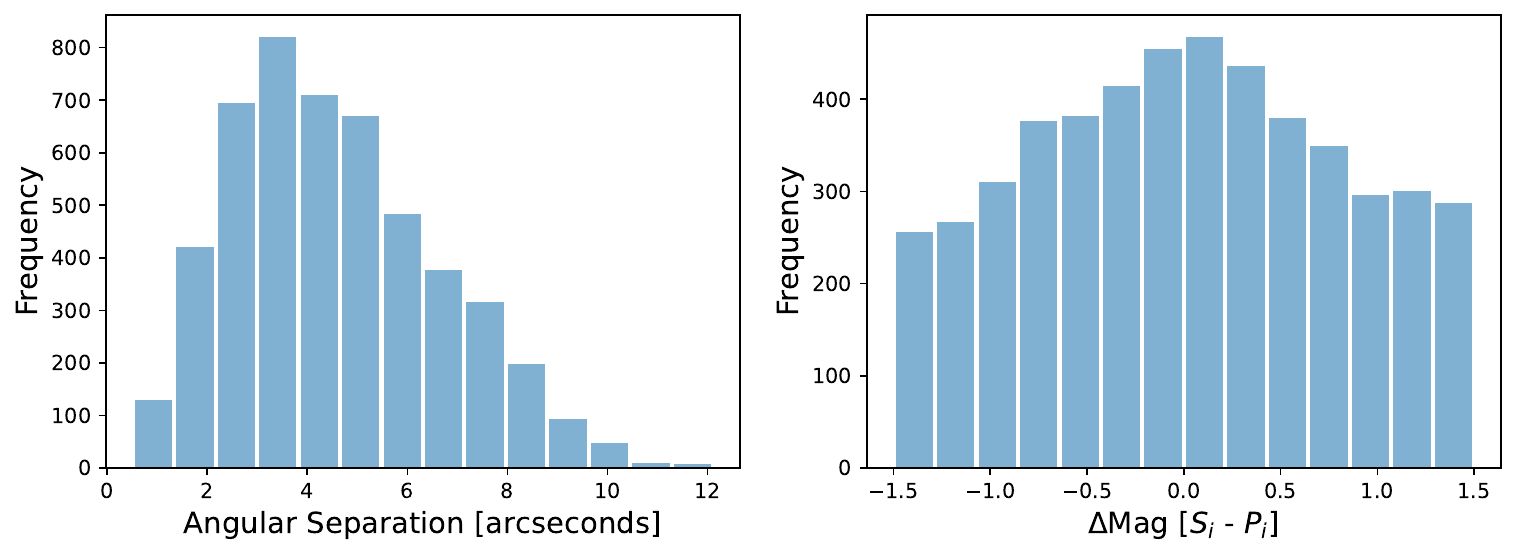}
	\label{fig:csodist}
	\caption{CSO Cutout Distributions.}
\end{figure}

Next, to determine the magnitude difference between the two satellites we use the following:

\begin{equation}
	\text{Magnitude Difference} = S_i - P_i,
\end{equation}

where $S_i$ is the i-band magnitude of the secondary satellite, and $P_i$ is the i-band magnitude of the primary satellite. The distribution of these quantaties across our simulated dataset is shown in Fig. \ref{fig:csodist}. Please note that the reason for the plateau towards $\Delta Mag = \pm1.5$ is due to the random generator selecting a number that then makes the secondary satellite have a magnitude that falls outside of the magnitude limit bounds set up in the simulation (given in Table \ref{tab:simparams}). When this happens, the secondary satellite gets a magnitude equal to that of the closest value, which eneds up being the high or low boundary for satellite magnitude in the given simulation.

\section{Methods}

To compare how \texttt{MuyGPyS} performs on our simulated dataset, we compare it against a logistic regression model and a neural network (NN) model. Reference \cite{Buchanan:2022} demonstrated the utility of using \texttt{MuyGPyS} on galaxy blends, which is very similar to the CSO vs. single satellite classification problem, and they simulate nearly the same sized cutouts using similar methods as presented in this work. Because the galaxy blend and satellite ``blend" problems have such overlap, we borrow many of the same parameters and values from that previous analysis \cite{Buchanan:2022}.  

The following subsections cover the three models used, and include tables with each hyperparameter variable that we fix or do sweeps over. For the neural network we directly replicate the model from \cite{Buchanan:2022} (please see reference \cite{Buchanan:2022} for a more thorough explanation of parameter choices), while the other two models tested here use a mixture of replicated values, and performing a hyperparameter sweep to determine our own best fitting values. For each hyperparameter sweep, we take the average over many runs, and set that as the best fitting parameter. The best fit results for each sweep are within tables in the subsections below. 

For each classification method used in this work, we use a training and testing data split (resulting from a hyperparameter sweep) of:
\begin{equation}
	\text{Train/Test Split} = 80/20
\end{equation} 

\subsection{Logistic Regression Model}

We choose a logistic regression model that enables and supports cross-validation techniques. Our best fitting parameters for this model are reported in \ref{tab:logreg}. We choose the L2 (Ridge Regression) penalty term, which adds a squared term as a penalty term to the loss function.

Before running the logistic regression model, we do Principle Component Analysis (PCA) on the flattened 1-D image pixel value arrays. We did a sweep over the number of components for the PCA and determined 21 components to be the best number for the PCA. 

\begin{table}[h!]
	\begin{center}
		\caption{Best Fitting Logistic Regression Parameters.}
		\begin{tabular}{|c|c|} 
			\hline
			\textbf{Parameter} & \textbf{Value} \\ 
			\hline
			Number of cross-validation folds & 2 \\ 
			Penalty term & L2 (Ridge Regression) \\
			Number of iterations for optimization & 10,000 \\
			\hline
		\end{tabular}
		\label{tab:logreg}
	\end{center}
\end{table}

\subsection{Neural Network (NN) Model}

We directly replicate the sequential neural network model used in previous works \cite{Buchanan:2022}, aside from the loss function, which we change from a binary cross-entropy loss term to a sparse categorical cross-entropy loss term, as it performed better for our specific analysis. The neural network layers used in our sequential model are listed in Table \ref{tab:nnlayers}, and all other model parameters are listed in Table \ref{tab:nn}. We do not do a PCA on the images for this model, so the images retain their original 24$\times$24 pixel shape. 

\begin{table}[h!]
	\begin{center}
		\caption{Sequential Neural Network Model.}
		\begin{tabular}{|c|c|c|} 
			\hline
			\textbf{Layer Number} & \textbf{Layer} & \textbf{Parameters} \\ 
			\hline
			\hline
			1 & 2D Convolutional Layer & number of channels: 128\\
			& & kernel size: (3, 3)\\
			& & cross-correlation stride: (1, 1)\\
			& & activation function = ReLu \\
			\hline 
			\hline
			2 & 2D Max Pooling Layer & pooling window: (2, 2) \\
			& & pool window stride: (2, 2) \\
			\hline
			\hline
			3 & 2D Convolutional Layer & number of channels: 64\\
			& & kernel size: (3, 3)\\
			& & cross-correlation stride: (1, 1)\\
			& & activation function = ReLu \\
			\hline
			\hline
			4 & 2D Max Pooling Layer & pooling window: (2, 2) \\
			& & pool window stride: (2, 2) \\
			\hline
			\hline
			5 & Flatten Layer & \\
			\hline
			\hline
			6 & Dense Layer & output space dimensionality: 800 \\
			& & activation function: ReLu \\
			\hline
			\hline
			7 & Dropout Layer & Rate: 0.2 \\
			\hline
			\hline
			8 & Dense Layer & output space dimensionality: 400 \\
			& & activation function: ReLu \\
			\hline
			\hline
			9 & Dropout Layer & Rate: 0.2 \\
			\hline
			\hline
			10 & Dense Layer & output space dimensionality: 200 \\
			& & activation function: ReLu \\
			\hline
			\hline
			11 & Dense Layer & output space dimensionality: 2 \\
			& & activation function: softmax \\
			\hline
		\end{tabular}
		\label{tab:nnlayers}
	\end{center}
\end{table}

\begin{table}[h!]
	\begin{center}
		\caption{NN Model Values.}
		\begin{tabular}{|c|c|} 
			\hline
			\textbf{Parameter} & \textbf{Value} \\ 
			\hline
			Number of epochs & 15 \\ 
			Batch size & 200 \\ 
			Loss function & Sparse Categorical Crossentropy\\
			Optimizer & Adam \\
			\hline
		\end{tabular}
		\label{tab:nn}
	\end{center}
\end{table}

\subsection{MuyGPyS}

We choose to use a Mat{\`{e}}rn kernel in our \texttt{MuyGPyS} model, which transforms crosswise tensors into cross-covariance tensors, and pairwise tensors into covariance tensors. This kernel enables the user to specify distortion model parameters, including $nu$, which is the smoothness of the resulting functions, and the distance function to be used (which includes the length scale). The values we use in this work are listed in Table \ref{tab:muygpys}. The MuyGPyS methods are used on the PCA reduced data.

\begin{table}[h!]
	\begin{center}
		\caption{Best Fitting \texttt{MuyGPyS} Parameter Values.}
		\begin{tabular}{|c|c|c|} 
			\hline
			\textbf{Parameter} & \textbf{Value} \\ 
			\hline
			Number of nearest neighbors & 28 \\ 
			Kernel & Mat{\`{e}}rn \\
			nu & 10 \\
			Distortion & Isotropic \\
			Length scale & 20 \\
			\hline
		\end{tabular}
		\label{tab:muygpys}
	\end{center}
\end{table}

\section{Results}

\begin{figure}[h!]
	\centering
	\includegraphics[width=0.85\textwidth]{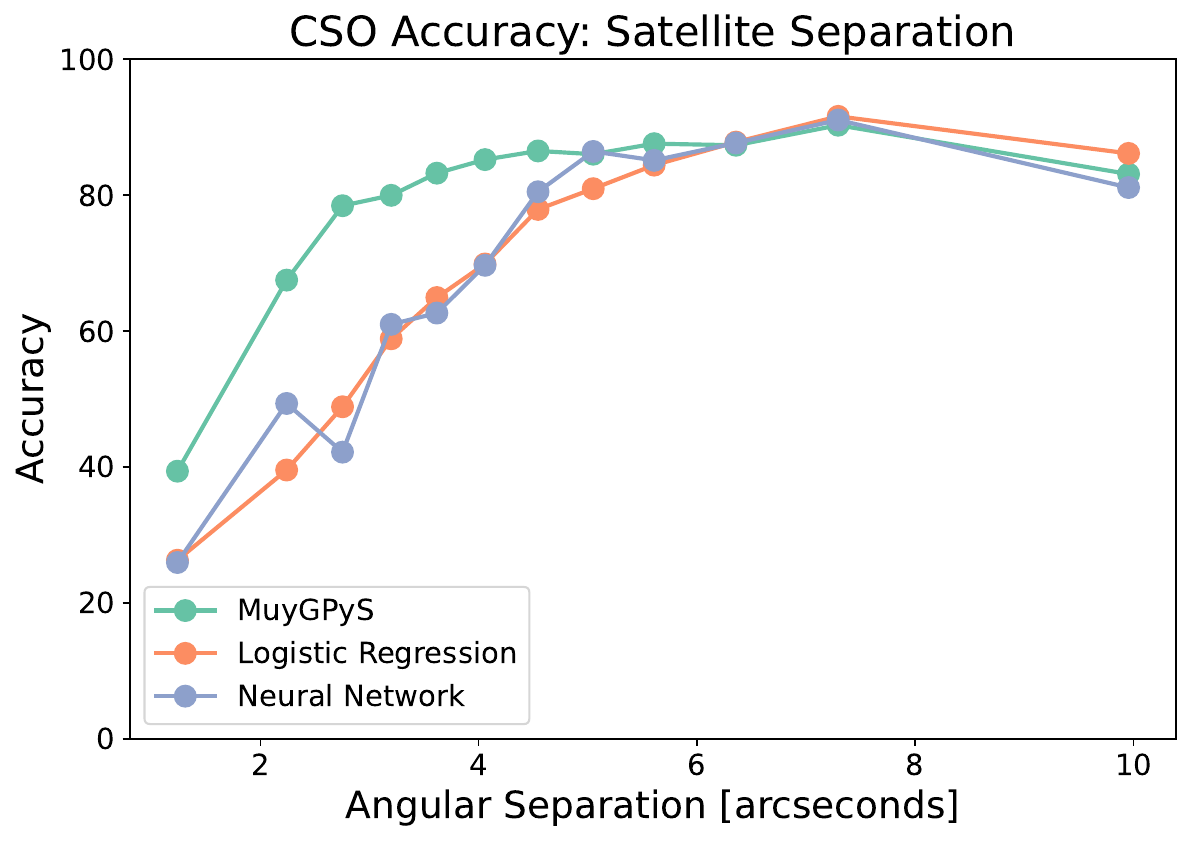}
	\caption{CSO Classification Accuracy as a Function of Angular Separation.}
	\label{fig:sep}
\end{figure}

\begin{figure}[h!]
	\centering
	\includegraphics[width=0.85\textwidth]{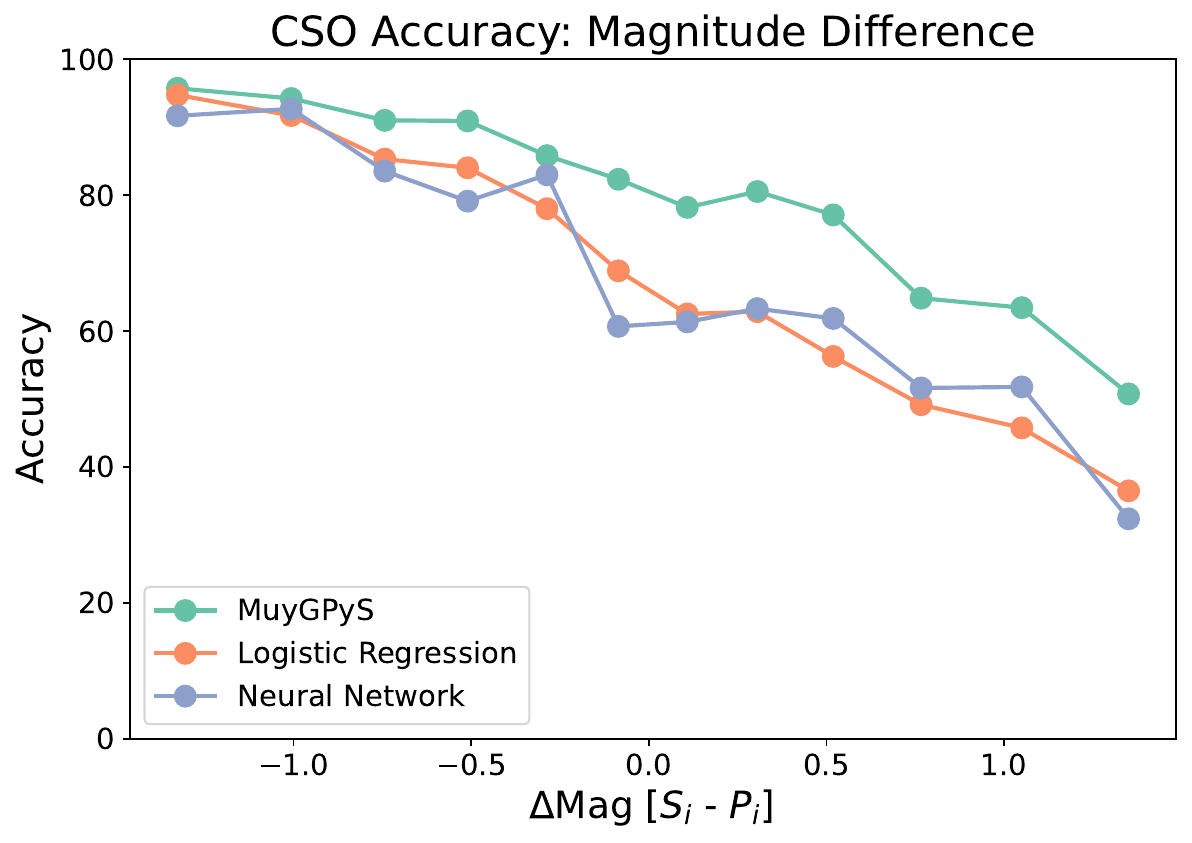}
	\caption{CSO Classification Accuracy as a Function of Magnitude Difference.}
	\label{fig:magdiff}
\end{figure}

\begin{figure}[h!]
	\centering
	\includegraphics[width=0.9\textwidth]{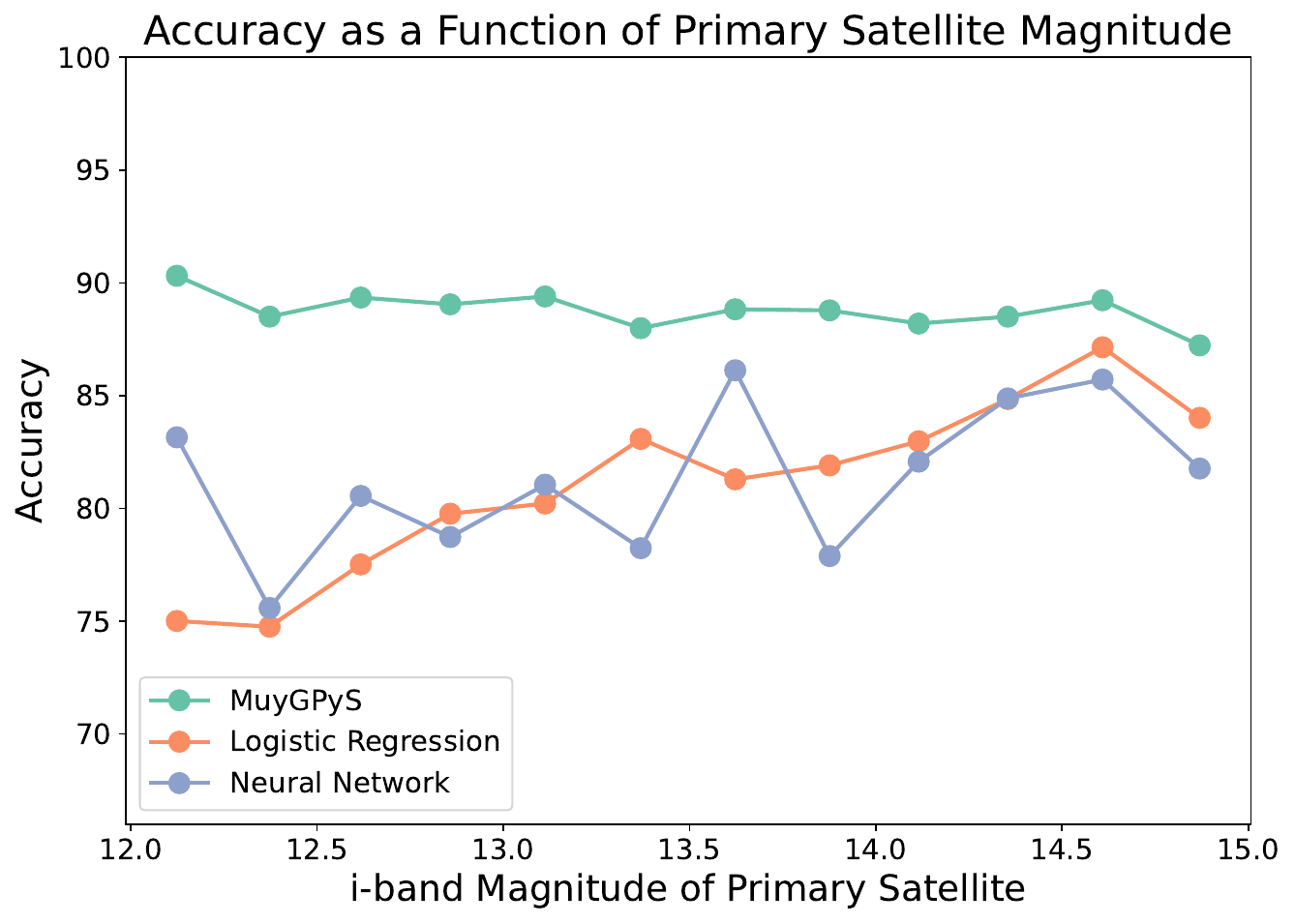}
	\caption{Classification Accuracy as a Function of Primary Satellite Magnitude.}
	\label{fig:mag}
\end{figure}

Fig. \ref{fig:sep} and Fig. \ref{fig:magdiff} show the classification accuracy for the CSO class, as a function of the angular seaparation between satellites, and the magnitude difference between the primary and secondary satellite pair. Fig. \ref{fig:mag} shows the classification accuracy as a function of the magnitude of the primary satellite (meaning it applies to both the singular and CSO class). All results shown are averaged over 100 runs of each model. Each distribution of values (the x-axis on each plot) is split into 12 bins, in which there is an equal number of data points in each bin. The accuracy is plotted on the y-axis at the point on the x-axis that corresponds to the average of that bin range. 

Fig. \ref{fig:sep}, Fig. \ref{fig:magdiff}, and Fig. \ref{fig:mag} illustrate that \texttt{MuyGPyS} outperforms the logistic regression and neural network models. This is true for every bin across each accuracy comparison, except where it starts to break down as a function of angular separation increasing. When the angular separation between two satellites becomes $>8 arcseconds$, the logistic regression, neural network, and GP model all degrade to essentially the same performace. This is due to the fact that at this separation, the images start to get so far separated that they begin to get classified as a single satellite again (see ``CSO 12" in Fig. \ref{fig:csosats} for a good example of an average signal-to-noise simulated cutout near this classification degredation boundary).

\begin{figure}[h!]
	\vspace{5em}
	\centering
	\includegraphics[width=0.87\textwidth]{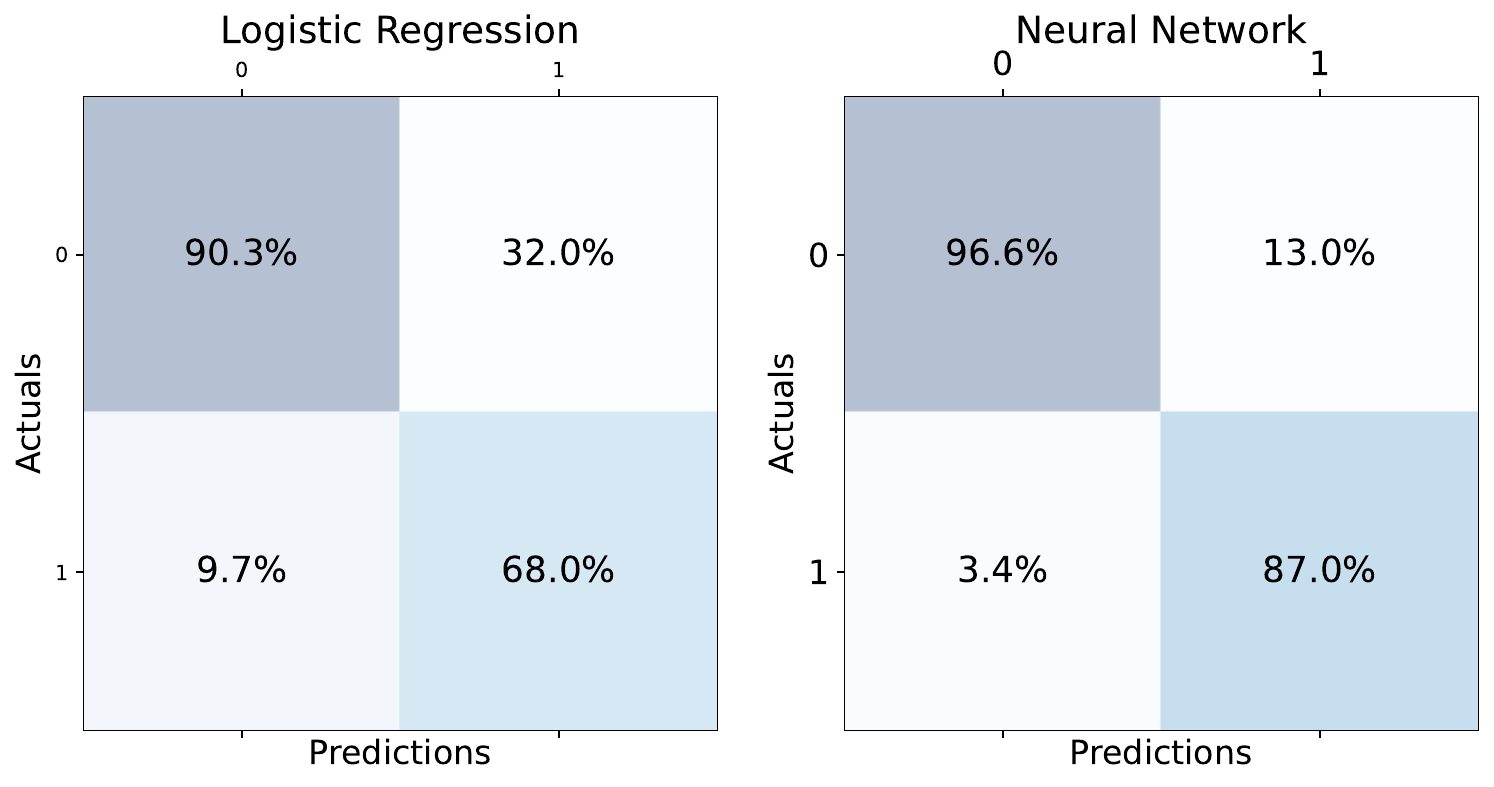}
	\caption{Confusion Matrices.}
	\label{fig:conf}
\end{figure}

\texttt{MuyGPyS} performs especially well when correctly classifying two satellites as a CSO, which is the most stressing case for SDA operators and decision makers. For the comparison between accuracy as a function of angular separation, CSO magnitude difference, and primary satellite magnitude, the neural network and logistic regression methods appear to perform about the same (within the uncertainty in the models). However, taking a closer look at the confusion matrices in Fig. \ref{fig:conf} reveals that the neural network does better at correctly classifying the single satellite and CSO cases, while also minimizing the number of false positive and false negatives, making it superior to the logistic regression model. 

Fig. \ref{fig:magdiff} additionally demonstrates that \texttt{MuyGPyS} outperforms the other two methods when it comes to magnitude differences between the two satellites in proximity. When the primary satellite is a higher magnitude than the secondary satellite, the signal from the primary satellite can dominate, making it hard to detect a second object close by. \texttt{MuyGPyS} performs well in this region, while the neural network and logistic regression model degrade at a steeper rate $\Delta Mag$. 

\section{Conclusion}

Quickly and accurately identifying whether an RPO maneuver is happening between two satellites in close proximity is critical for the safety of space missions and human spaceflight. When two satellites are close together along the line-of-sight they can fall within the PSF full-width half-max of the optical system, causing them to appear as a single object, or ``blended". If the satellites are close enough, one satellite is brighter than the other, or there is poor signal-to-noise, seeing, etc., the human eye cannot distinguish one vs. two satellites given ground-based COTS electro-optical observations. Throughout this work we have demonstrated how we use LLNL-developed SDA simulation packages to simulate images of single satellites and CSOs accross a range of observing conditions, and sensor and satellite parameters. We then describe our data processing techniques and show the utility of classifying these cutouts using GP methods.

\texttt{MuyGPyS} is a fast, flexible, and accurate open-source python package that can be utilized for SDA image classification tasks. We have demonstrated that \texttt{MuyGPyS} is capable of outperforming other traditional machine learning methods when it comes to more challenging areas of the problem parameter space. That is, \texttt{MuyGPyS} can classify CSOs under stressing conditions, such as when satellites are very close in proximity, or there is a large magnitude difference (with the primary satellite having the higher magnitude) between the primary and secondary satellites. Classifying images in these regions is crucial for the future of SDA, and GP methods will likely be the solution for fast and reliable answers.

\section{Acknowledgements}

This work was performed under the auspices of the U.S. Department of Energy by Lawrence Livermore National Laboratory under Contract DE-AC52-07NA27344 and was supported by the LLNL LDRD Program under Project Number 22-ERD-054. We would like to thank James Buchanan, Daigo Kobayashi, and Josh Meyers for thoughtful and meaningful technical discussions. LLNL-PROC-854140.

\section{References}

\bibliographystyle{unsrt}

\begingroup
\renewcommand{\section}[2]{}%
\bibliography{references}
\endgroup

\end{document}